\documentclass[conference]{IEEEtran}
\IEEEoverridecommandlockouts
\usepackage{cite}
\usepackage{amsmath,amssymb,amsfonts}
\usepackage{algorithmic}
\usepackage{graphicx}
\usepackage{textcomp}
\usepackage{xcolor}

\usepackage{multirow}
\usepackage{subfigure}

\usepackage{xcolor}
\usepackage{array} 

\def\BibTeX{{\rm B\kern-.05em{\sc i\kern-.025em b}\kern-.08em
    T\kern-.1667em\lower.7ex\hbox{E}\kern-.125emX}}
\begin{document}

\title{A Practical Gated Recurrent Transformer Network Incorporating Multiple Fusions for Video Denoising\\
%{\footnotesize \textsuperscript{*}Note: Sub-titles are not captured for https://ieeexplore.ieee.org  and
}

\author{\IEEEauthorblockN{1\textsuperscript{st} Kai Guo}
	\IEEEauthorblockA{\textit{Samsung Electronics} \\
		 Hwaseong-si, Korea \\
		visionkai@gmail.com}
	\and
	\IEEEauthorblockN{2\textsuperscript{nd} Seungwon Choi}
	\IEEEauthorblockA{\textit{Samsung Electronics} \\
		 Hwaseong-si, Korea \\
         sw5190.choi@samsung.com}
	\and
	\IEEEauthorblockN{3\textsuperscript{rd} Jongseong Choi}
	\IEEEauthorblockA{\textit{Samsung Electronics} \\
	Hwaseong-si, Korea \\
	jongseong.choi@samsung.com}
	\and
	\IEEEauthorblockN{4\textsuperscript{th} Lae-Hoon Kim}
	\IEEEauthorblockA{\textit{Samsung Electronics} \\
	Hwaseong-si, Korea \\
	laehoon.kim@samsung.com}
}
%\vspace{-0.5cm}

%\vspace{-1.5cm}

%\author{\IEEEauthorblockN{1\textsuperscript{st} Kai Guo}
%\and
%\and
%\and
%\and
%\and

\maketitle

\begin{abstract}
State-of-the-art (SOTA) video denoising methods employ multi-frame simultaneous denoising mechanisms, resulting in significant delays (e.g., 16 frames), making them impractical for real-time cameras.
To overcome this limitation, we propose a multi-fusion gated recurrent Transformer network (GRTN) that achieves SOTA denoising performance with only a single-frame delay.
Specifically, the spatial denoising module extracts features from the current frame, while the reset gate selects relevant information from the previous frame and fuses it with current frame features via the temporal denoising module.
The update gate then further blends this result with the previous frame features, and the reconstruction module integrates it with the current frame.
To robustly compute attention for noisy features, we propose a residual simplified Swin Transformer with Euclidean distance (RSSTE) in the spatial and temporal denoising modules. Comparative objective and subjective results show that our GRTN achieves denoising performance comparable to SOTA multi-frame delay networks, with only a single-frame delay.
\end{abstract}


\vspace{-0.5cm}
\section{Introduction}
\vspace{-0.1cm}

Despite advances in imaging sensors, shot noise and readout noise continue to significantly degrade image quality \cite{Foi08:TIP, Hasinoff10:CVPR, Brooks19:CVPR}. 
Deep learning-based video denoising methods have achieved state-of-the-art (SOTA) performance \cite{Tassano19:ICIP, Tassano20:CVPR, Yue20:CVPR, Vaksman21:CVPR, Liang24:TIP, Liang22:NIPS}, with the most effective techniques denoising multiple frames simultaneously \cite{Liang24:TIP, Liang22:NIPS}. However, these methods rely on future frames (e.g., 15 frames), introducing significant delays (e.g., 16 frames), making them unsuitable for real-time camera applications.

%, making denoising a crucial function in image signal processors (ISPs) within imaging devices.

\begin{figure}
	%\vspace{-0.0cm}
	\centerline{\includegraphics[width=3.0in]{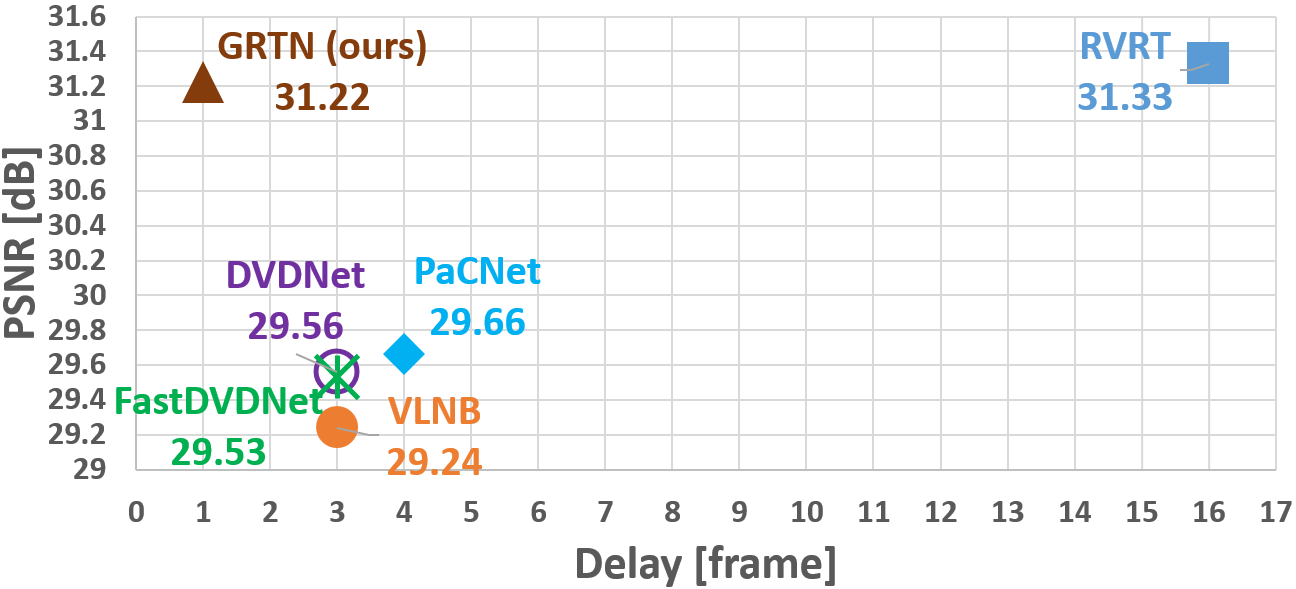}}
	\vspace{-0.2cm}
	\caption{The proposed GRTN offers denoising performance comparable to SOTA multi-frame delay networks, but with only a single-frame delay. In contrast, VLNB\cite{Arias18:JMIV}, DVDNet\cite{Tassano19:ICIP} and FastDVDNet\cite{Tassano20:CVPR} have a 3-frame delay, while PaCNet\cite{Vaksman21:CVPR} and RVRT\cite{Liang22:NIPS} have 4- and 16-frame delays, respectively. The Set8 dataset \cite{Tassano19:ICIP} with Gaussian noise ($\sigma=50$) is used in this evaluation.}
	\label{fig:PSNR_delay}
	\vspace{-0.5cm}
\end{figure}

In this paper, we propose a multi-fusion gated recurrent Transformer network (GRTN) that achieves SOTA denoising performance with only a single-frame delay, as shown in Fig. \ref{fig:PSNR_delay}.
Specifically, the spatial denoising module first extracts features from the current frame. The reset gate selects relevant information from the previous frame, which is fused with the current frame features via the temporal denoising module. The update gate then blends this fusion result with previous frame features. Finally, the reconstruction module integrates the blended result with the current frame features.
For both the spatial and temporal denoising modules, we propose a residual simplified Swin Transformer with Euclidean distance (RSSTE), which offers greater robustness in calculating attention for noisy features and enhances the preservation of image details. Based on both subjective and objective experimental comparisons, the proposed network achieves performance comparable to SOTA multi-frame delay networks, while only requiring a single-frame delay.

The contributions of this work are summarized as follows:
\vspace{-0.1cm}
\begin{itemize}
	
	\item
We propose a gated recurrent Transformer network capable of performing multiple rounds of feature selection and fusion. 
The reset and update gates select relevant information from the previous frame, and integrate it with the current frame through temporal denoising and blending, respectively. The reconstruction module then further fuses the blended output with the current frame.

	\item
	We introduce the RSSTE Transformer, which incorporates Euclidean distance-based attention to enhance robustness in noisy conditions. Additionally, a constraint is applied to maximize orthogonality in the Transformer weights, promoting the learning of more independent features.
	
\end{itemize}

%We propose a gated recurrent transformer network capable of multiple rounds of feature selection and fusion. The reset gate selects relevant information from the previous frame, which is integrated with the current frames features through a temporal denoising module. The update gate blends this result with previous frame features, and the reconstruction module fuses the blended output with the current frames features.

\begin{figure*}
	%\vspace{-0.0cm}
	\centerline{\includegraphics[width=5.7in]{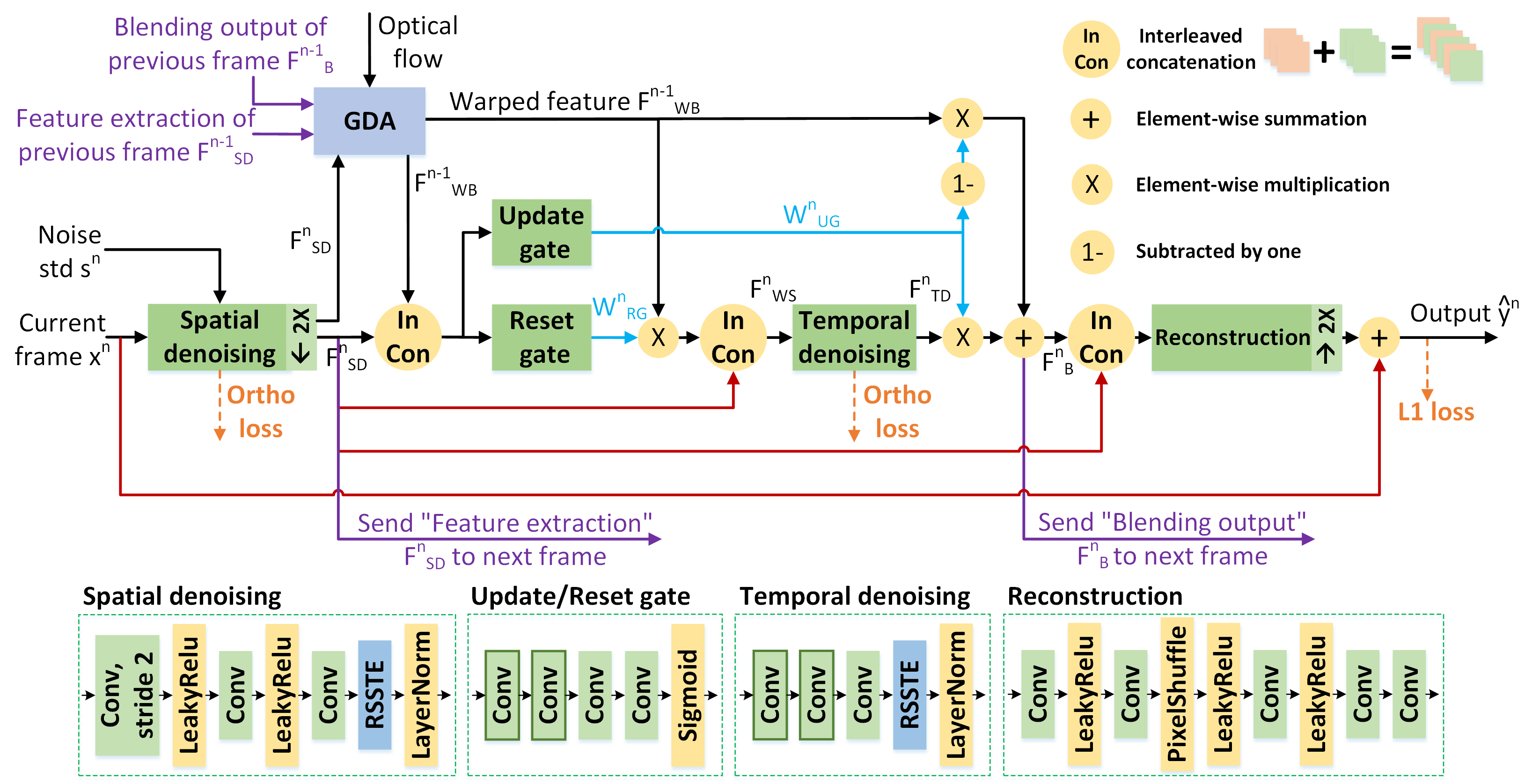}}
	\vspace{-0.3cm}
	\caption{The detailed network architecture of the proposed GRTN. GDA refers to guided deformable alignment \cite{Liang22:NIPS}.}
	\label{fig:architecture}
	\vspace{-0.5cm}
\end{figure*}

\vspace{-0.2cm}
\section{Related works}
\vspace{-0.1cm}

Classical signal processing methods \cite{Buades05:CVPR, Dabov07:TIP} and deep learning techniques \cite{Santhanam17:CVPR, Liu18:CVPR, Zhang17:TIP, Zhang18:TIP, Chang20:ECCV, Davy19:ICIP, Vogels18:ACMGraphics, Mildenhall18:CVPR} reduce noise by fusing information from similar textures. Recently, Transformer \cite{Vaswani17:NIPS, Liu21:ICCV} based networks \cite{Liang21:ICCVW, Liang24:TIP, Liang22:NIPS} have demonstrated convincing denoising performance.
Video denoising generally fuses information from similar textures across temporal and spatial domains \cite{Claus19:CVPRW, Wang19:CVPRW, Maggioni12:TIP, Maggioni21:CVPR, Tassano19:ICIP, Tassano20:CVPR, Vaksman21:CVPR, Liang24:TIP, Liang22:NIPS} and can be classified into sliding window-based, recurrent, and multi-frame simultaneous processing (MFSP) methods.
Sliding window methods \cite{Tassano19:ICIP, Tassano20:CVPR, Vaksman21:CVPR} denoise each frame by incorporating past and future frames, the resulting delays make them unsuitable for real-time applications.
Recurrent methods \cite{Fuoli19:ICCVW, Godard18:ECCV, Maggioni21:CVPR} leverage RNNs \cite{Rumelhart86:Nature} to integrate features from past frames, but they underutilize temporal redundancy, leading to performance below MFSP techniques.
MFSP methods \cite{Liang24:TIP, Liang22:NIPS} that process multiple frames simultaneously deliver superior denoising but introduce significant delays (e.g., 16 frames), making them impractical for real-time cameras.

\vspace{-0.15cm}
\section{Method}
\vspace{-0.1cm}

\subsection{Network Architecture}
%\vspace{-0.1cm}
%The architecture of the proposed GRTN is shown in Fig. \ref{fig:architecture}.
%Key components, including spatial denoising, reset and update gates, alpha blending, and reconstruction modules, enable multi-level fusion of features from the current and previous frames.
%The design of the reset and update gates is inspired by GRU and LSTM models \cite{Hochreiter97:NC, Cho14:SSST}.

The architecture of the proposed GRTN is shown in Fig. \ref{fig:architecture}.
The design of the reset and update gates is inspired by GRU \cite{Cho14:SSST} and LSTM models \cite{Hochreiter97:NC}.

\textbf{Spatial denoising}

The spatial denoising module comprises three convolutional layers with Leaky ReLU activations, an RSSTE module, and a normalization layer. The first convolutional layer includes $2\times$ downsampling. 
This module performs spatial denoising on the current frame $x^n$ based on its noise standard deviation $s^n$, while extracting high-dimensional features.
The process is defined as:
\begin{equation}
\vspace{-0.1cm}
\label{eq:spatial_denoising}
F^{n}_{SD} = H_{SD}(Con(x^n, s^n))
\vspace{-0.1cm}
\end{equation}
where $Con(\cdot)$ denotes concatenation, and $n$ is the frame index.
The output $F^{n}_{SD}$ is also used as the input for the next frame.

\textbf{Reset gate}

The blended features $F^{n-1}_{B}$ from the previous frame are warped using Guided Deformable Attention (GDA)\cite{Liang22:NIPS}, yielding $F^{n-1}_{WB}$. 
$F^{n-1}_{WB}$ and $F^{n}_{SD}$ are then interleaved and processed through a reset gate, which produces a weight $W^{n}_{RG}$ representing their similarity. 
Interleaving features from the same channel enables more precise and effective comparison.

The reset gate consists of four convolutional layers with a sigmoid activation function.
The first two layers use grouped convolutions (grouped by 2) to efficiently compare same-position channels. The reset gate is defined as: 
\begin{equation}
\vspace{-0.1cm}
\label{eq:resetGate}
W^{n}_{RG} = H_{RG}(InCon(F^{n}_{SD}, F^{n-1}_{WB}))
\vspace{-0.1cm}
\end{equation}
where $InCon(\cdot)$ denotes interleaved concatenation.
Next, the weight $W^{n}_{RG}$ is multiplied by $F^{n-1}_{WB}$ to extract relevant information, which is interleaved with $F^{n}_{SD}$ to produce $F^{n}_{WS}$:
\begin{equation}
\vspace{-0.1cm}
\label{eq:TD_input}
F^{n}_{WS} = InCon(W^{n}_{RG}{{\odot}}F^{n-1}_{WB}, F^{n}_{SD})
\vspace{-0.1cm}
\end{equation}
where $\odot$ represents element-wise multiplication.

\textbf{Temporal denoising}

The temporal denoising module fuses the interleaved concatenated feature $F^{n}_{WS}$ in the temporal domain. It comprises three convolutional layers, an RSSTE module, and a normalization layer. The first two convolutional layers use grouped convolutions (grouped by 2 and half the input channels, respectively) to efficiently fuse same-position channels. This can be expressed as: 
\begin{equation}
\vspace{-0.1cm}
\label{eq:spatial_denoising}
F^{n}_{TD} = H_{TD}(F^{n}_{WS})
\vspace{-0.1cm}
\end{equation}

\textbf{Update gate and alpha blending}

The update gate has the same structure as the reset gate, with inputs formed by interleaving $F^{n}_{SD}$ and $F^{n-1}_{WB}$. 
It is defined as:
\begin{equation}
\vspace{-0.1cm}
\label{eq:updateGate}
W^{n}_{UG} = H_{UG}(InCon(F^{n}_{SD}, F^{n-1}_{WB}))
\vspace{-0.1cm}
\end{equation}

The output weight $W^{n}_{UG}$ blends $F^{n}_{TD}$ and $F^{n-1}_{WB}$, effectively selecting the optimal components of $F^{n}_{TD}$ and the complementary elements from $F^{n-1}_{WB}$. This blending is defined as: 
\begin{equation}
\vspace{-0.1cm}
\label{eq:alpha_blending}
F^{n}_{B} = W^{n}_{UG}{\odot}F^{n}_{TD} + (1-W^{n}_{UG}){\odot}F^{n-1}_{WB}
\vspace{-0.1cm}
\end{equation}
The output $F^{n}_{B}$ also serves as input for the next frame.

\begin{figure}[htp]
	
	%\vspace{-0.0cm}
	\begin{minipage}[b]{0.3\linewidth}
		%\centering
		\centerline{\includegraphics[width= 0.9in]{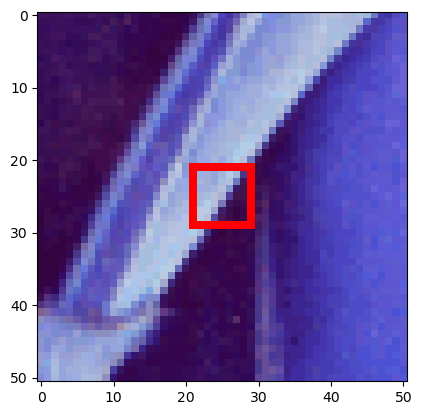}}
		\vspace{-0.05cm}
		\centerline{(a)}\medskip
	\end{minipage}
	\hfill
	\begin{minipage}[b]{0.3\linewidth}
		%\centering
		\centerline{\includegraphics[width= 1.1in]{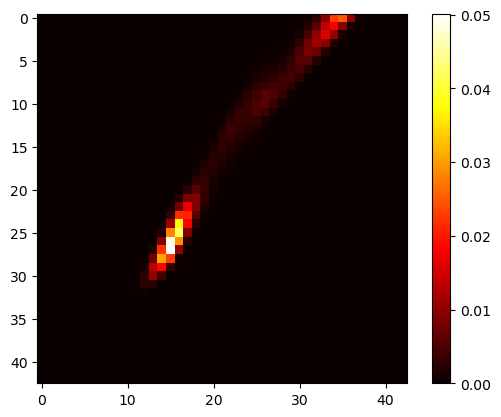}}
		\vspace{-0.05cm}
		\centerline{(b)}\medskip
	\end{minipage}
	\hfill
	%\hspace{1.46cm}
	\begin{minipage}[b]{0.3\linewidth}
		%\centering
		\centerline{\includegraphics[width= 1.13in]{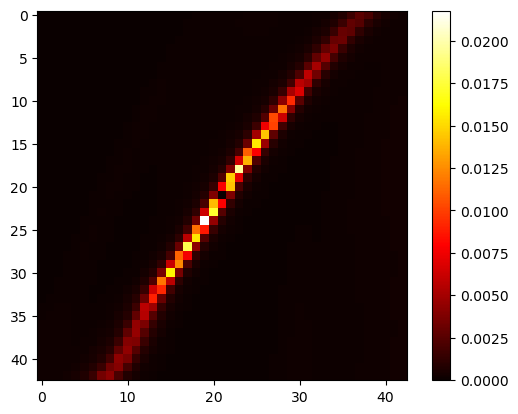}}
		\vspace{-0.05cm}
		\centerline{(c)}\medskip
	\end{minipage}
	
	\vspace{-0.2cm}
	\begin{minipage}[b]{0.3\linewidth}
		%\centering
		\centerline{\includegraphics[width= 0.9in]{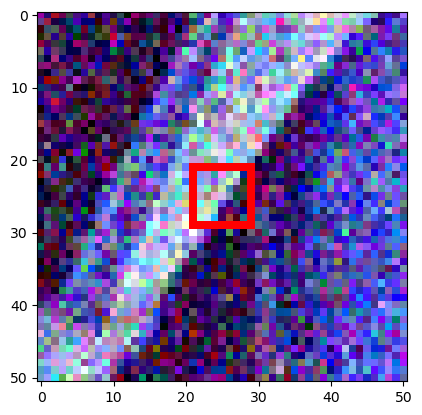}}
		\vspace{-0.05cm}
		\centerline{(d)}\medskip
	\end{minipage}
	\hfill
	\begin{minipage}[b]{0.3\linewidth}
		%\centering
		\centerline{\includegraphics[width= 1.1in]{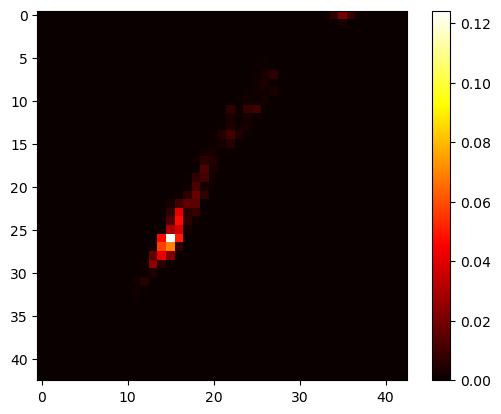}}
		\vspace{-0.05cm}
		\centerline{(e)}\medskip
	\end{minipage}
	\hfill
	%\hspace{1.46cm}
	\begin{minipage}[b]{0.3\linewidth}
		%\centering
		\centerline{\includegraphics[width= 1.13in]{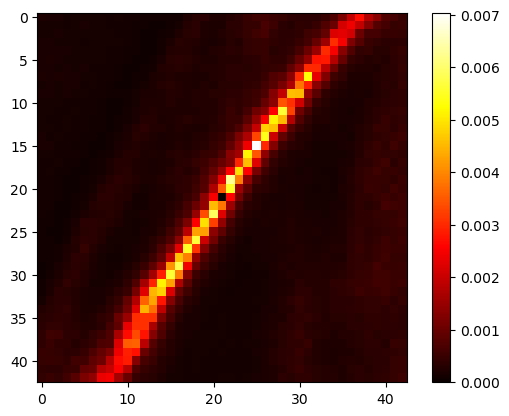}}
		\vspace{-0.05cm}
		\centerline{(f)}\medskip
	\end{minipage}
	\vspace{-0.4cm}
	\caption {Comparison of attention maps using dot product and Euclidean distance. (a) and (d) show a noise-free image (cropped from Lenna) and the same image with Gaussian noise ($\sigma=50$), respectively, with the central $9{\times}9$ patch highlighted in red. (b) and (e) display dot product-based attention maps for the central patch, calculated from (a) and (d). (c) and (f) show the corresponding Euclidean distance-based attention maps.} \label{fig:attention_map}
	\vspace{-0.6cm}
\end{figure}

\textbf{Reconstruction}

The reconstruction module refines the blended features $F^{n}_{B}$ by fusing them with $F^{n}_{SD}$. 
It comprises five convolutional layers, a pixel shuffle layer, and three leaky ReLU layers, enabling feature fusion and upsampling to produce a residual image. The process is expressed as:
\begin{equation}
\vspace{-0.1cm}
\label{eq:reconstruction}
\hat{y}^{n}_{RC} = H_{RC}(InCon(F^{n}_{B}, F^{n}_{SD}))
\vspace{-0.1cm}
\end{equation}

The output $\hat{y}^{n}_{RC}$ is then added to the input noisy frame $x^n$ to generate the final result: 
\begin{equation}
\vspace{-0.1cm}
\label{eq:reconstruction}
\hat{y}^{n} = \hat{y}^{n}_{RC} + x^{n}
\vspace{-0.1cm}
\end{equation}

\textbf{Loss function}

The loss function is a weighted sum of the L1 loss and orthogonality loss $O$ from the RSSTE module. 
The L1 loss ensures the output closely matches the ground truth, while $O$ measures the correlation between weight vectors in each RSSTE layer. 
Minimizing $O$ increases the independence of these vectors, enabling RSSTE to learn the most representative features. The loss function is defined as:
\begin{equation}
\vspace{-0.1cm}
\label{eq:loss_function}
L = \frac{1}{N}\sum_{n}^{N}(|\hat{y}^n - y^n| + {\lambda}O^n)
\vspace{-0.1cm}
\end{equation}
where $\lambda$ is tuning weight, $y^n$ is the ground truth, and $n$ is the frame index. 
The orthogonality loss $O$ is defined as: 
\begin{equation}
\vspace{-0.1cm}
\label{eq:loss_ortho}
O = \frac{1}{K}\sum_{k}^{K}(\frac{\sum_{a,b}^{A,B}(C_{k}^{ab})-\sum_{a}^{A}(C_{k}^{aa})}{A(B-1)})
\vspace{-0.1cm}
\end{equation}
where $C_k$ denotes the covariance matrix of weights $W_k$ in the $k$-th linear layer of all RSSTEs, $a, b$ are element indices. 
$C_k$ is defined as $C_k = (W_k - \overline{W_k}) \times (W_k - \overline{W_k})^T$
with $\overline{W_k}$ as the row-wise mean of $W_k$. 
The loss $O$ is the average of the off-diagonal elements of $C_k$, averaged across all layers, quantifying the correlation between the row vectors.

%\begin{equation} 
%	C_k = (W_k - \overline{W_k}) \times (W_k - \overline{W_k})^T
%\end{equation} 

%\subsection{Residual Simplified Swin Transformer with Euclidean distance}

\vspace{-0.15cm}
\subsection{RSSTE}
\vspace{-0.1cm}

We propose the RSSTE Transformer model, which employs Euclidean distance-based attention to enhance accuracy and robustness in handling noisy features, outperforming the traditional dot product method. As shown in Fig. \ref{fig:attention_map}, we compare attention using dot product and Euclidean distance in both noise-free and noisy conditions. Fig. \ref{fig:attention_map}(b) and (c) depict attention on a noise-free image, while Fig. \ref{fig:attention_map}(e) and (f) show attention on a noisy image. Under noisy conditions, Euclidean distance-based attention more closely aligns with the noise-free attention than the dot product-based approach.

\begin{figure}
	%\vspace{-0.0cm}
	\centerline{\includegraphics[width=2.4in]{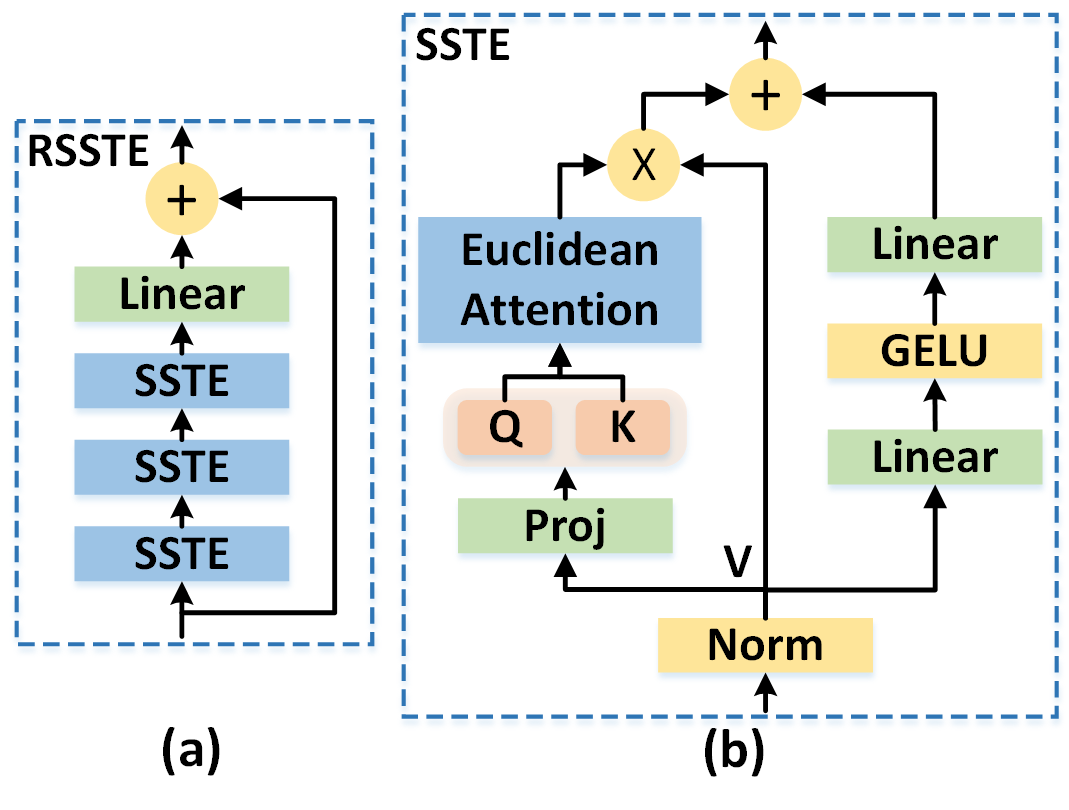}}
	\vspace{-0.3cm}
	\caption{(a) Residual simplified Swin Transformer with Euclidean attention (RSSTE). (b) Simplified Swin Transformer with Euclidean attention (SSTE).}
	\label{fig:RSSTE}
	\vspace{-0.5cm}
\end{figure}

%The structure of the RSSTE network consists of several layers of Simplified Swin Transformer with Euclidean distance (SSTE), a linear layer, and a residual connection, as shown in Fig. \ref{fig:RSSTE}(a).
%Assuming the input feature of RSSTE is $F^{n}_{0}$, 
%we first extract intermediate features $F^{n}_{1},F^{n}_{2},...,F^{n}_{J}$ through $J$ layers of SSTE:
%\begin{equation}
%	%\vspace{-0.1cm}
%	\label{eq:RSSTE_INTER}
%	F^{n}_{j} = H_{SSTE_{j}}(F^{n}_{j-1}),  j=1,2,...,J
%	%\vspace{-0.1cm}
%\end{equation}
%where $H_{SSTE_{j}}(\cdot)$ is the $j-th$ SSTE.
%Then the output of RSSTE is obtained through a linear layer and a residual connection:
%\begin{equation}
%	\vspace{-0.1cm}
%	\label{eq:RSSTE_OUPUT}
%	F^{n}_{out} = H_{LINEAR}(F^{n}_{J}) + F^{n}_{0}
%	\vspace{-0.1cm}
%\end{equation}
%where $H_{LINEAR}(\cdot)$ is the linear mapping layer within RSSTE.

The RSSTE network is composed of multiple simplified Swin Transformer with Euclidean distance (SSTE) layers, a linear layer, and a residual connection, as shown in Fig.\ref{fig:RSSTE}(a). 
Given the input feature $F^{n}_{0}$, intermediate features $F^{n}_{1},F^{n}_{2},...,F^{n}_{J}$ are extracted through $J$ SSTE layers:
\begin{equation}
	%\vspace{-0.1cm}
	\label{eq:RSSTE_INTER}
	F^{n}_{j} = H_{SSTE_{j}}(F^{n}_{j-1}),  j=1,2,...,J
	%\vspace{-0.1cm}
\end{equation}
where $H_{SSTE_{j}}(\cdot)$ represents the $j$-th SSTE layer. 
RSSTE output is computed via a linear layer with a residual connection:
\begin{equation}
	\vspace{-0.1cm}
	\label{eq:RSSTE_OUPUT}
	F^{n}_{out} = H_{LINEAR}(F^{n}_{J}) + F^{n}_{0}
	\vspace{-0.1cm}
\end{equation}
where $H_{LINEAR}(\cdot)$ denotes the linear mapping layer.

%SSTE first partitions the input with size $H{\times}W{\times}C$ into features of size $\frac{HW}{M^2}{\times}{M^2}{\times}C$, 
%where $M{\times}M$ is the size of the local window, and $\frac{HW}{M^2}$ is the number of windows.
%For a feature $V$ in a local window of size ${M^2}{\times}C$, 
%the query $Q$ and key $K$ are obtained through linear projection, as follows
%\begin{equation}
%	\vspace{-0.1cm}
%	\label{eq:SSTE_linearProj}
%	Q=V{P_Q}, K=V{P_K}
%	\vspace{-0.1cm}
%\end{equation}
%where $P_Q$ and $P_K$ are the projection matrices of the linear layers, 
%and $Q$ and $K$ have the same size as $V$, which is ${M^2}{\times}C$.
%Then, we calculate self-attention using the Euclidean distance between $Q$ and $K$
%\begin{equation}
%	%\vspace{-0.1cm}
%	\label{eq:SSTE_EUCLIDEAN}
%	Attention(Q,K,V)=SoftMax(-({\|{Q-K}\|_2}) + B)V
%	%\vspace{-0.1cm}
%\end{equation}
%where $B$ is the learnable relative positional encoding.
%Since $SoftMax(\cdot)$ uses the $exp(\cdot)$ function, the distance finally becomes $exp(-1{\times}\sqrt{({\|{Q-K}\|_2})}^2)$, 
%which means taking the negative square of the Euclidean distance followed by the exponential function. 
%The shorter this distance, the greater the resulting attention.
%We also perform the attention operation in parallel $h$ times and concatenate the resulting outputs, 
%which forms the multi-head self-attention (MSA).

SSTE first partitions the input of size $H{\times}W{\times}C$ into features of size $\frac{HW}{M^2}{\times}{M^2}{\times}C$, where $M{\times}M$ is the local window size, and $\frac{HW}{M^2}$ is the number of windows. For a feature $V$ in a local window of size ${M^2}{\times}C$, the query $Q$ and key $K$ are obtained via linear projection:
\begin{equation}
\vspace{-0.1cm}
\label{eq:SSTE_linearProj}
Q=V{P_Q}, K=V{P_K}
\vspace{-0.1cm}
\end{equation}
where $P_Q$ and $P_K$ are the projection matrices, and $Q$ and $K$ have the same size as $V$, i.e., ${M^2}{\times}C$. 
Self-attention is then computed using the Euclidean distance between $Q$ and $K$: 
\begin{equation}
%\vspace{-0.1cm}
\label{eq:SSTE_EUCLIDEAN}
Attention(Q,K,V)=SoftMax(-({\|{Q-K}\|_2}) + B)V
%\vspace{-0.1cm}
\end{equation}
 where $B$ is the learnable relative positional encoding. 
 The distance becomes $exp(-({\|{Q-K}\|_2}))$ after applying the exponential function, emphasizing greater attention for shorter distances. 
We perform this attention operation $h$ times in parallel and concatenate the outputs to form multi-head self-attention (MSA).

% The distance becomes $exp(-1{\times}\sqrt{({\|{Q-K}\|_2})}^2)$ after applying the exponential function, emphasizing greater attention for shorter distances. 
% We perform this attention operation in parallel $h$ times and concatenate the outputs to form multi-head self-attention (MSA).

\begin{figure*}[htbp]
	\centering
	\subfigure{
		\includegraphics[width=6.6in]{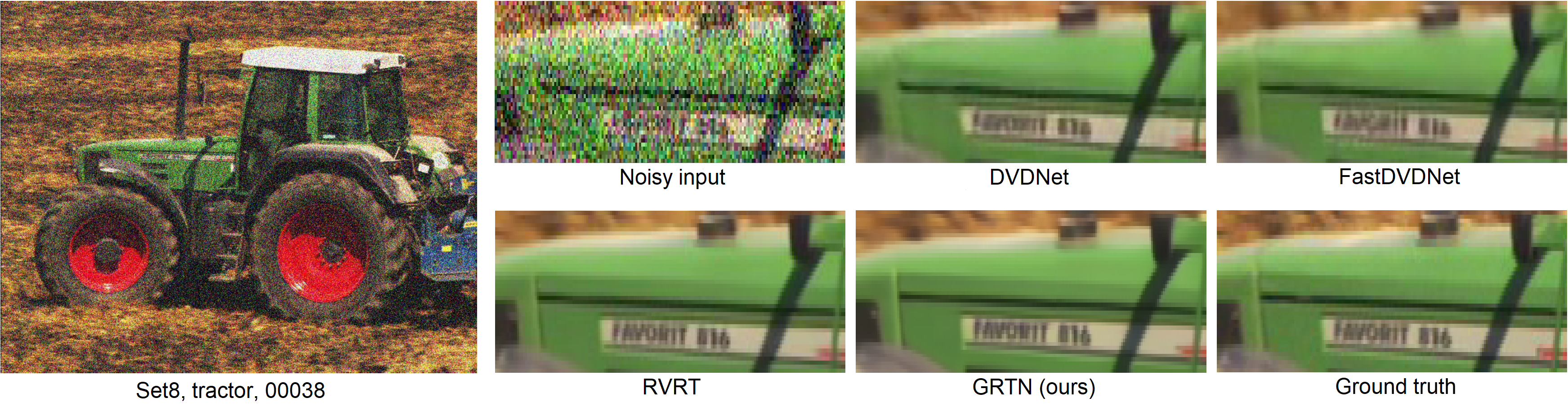}
		\label{fig:comp_set8_tractor}
	}
	%\vspace{0.1cm} % 控制上下图片间距
	\subfigure{
		\includegraphics[width=6.6in]{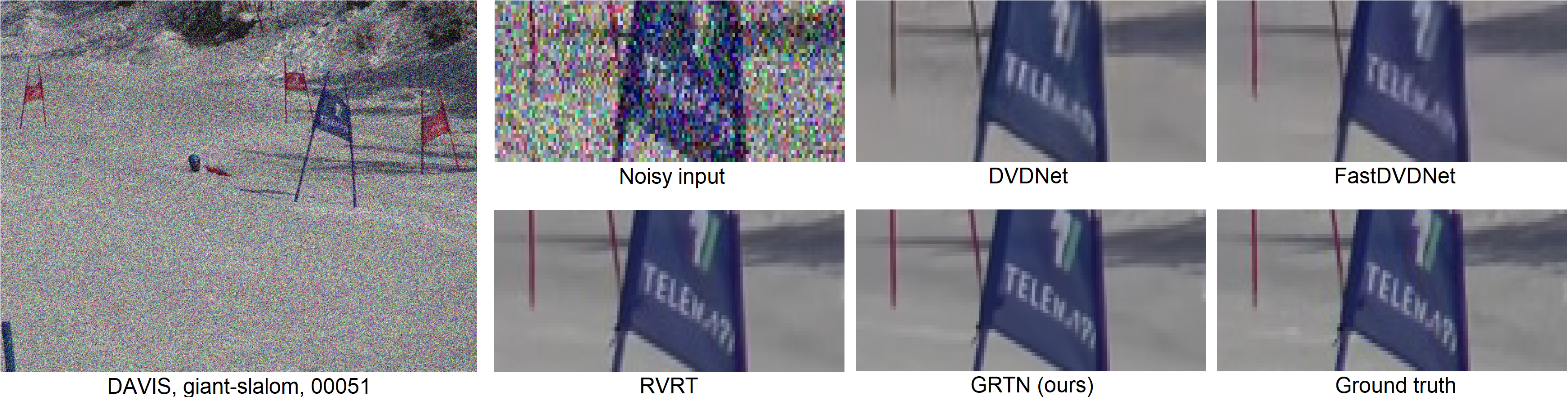}
		\label{fig:comp_davis_slalom}
	}
\vspace{-0.3cm}
	\caption{Video denoising comparison ($\sigma=50$) on Set8\cite{Tassano19:ICIP} and DAVIS\cite{Khoreva18:ACCV}.}
	\label{fig:comp_set8_davis}
\end{figure*}
%\vspace{-0.0cm}

%In SSTE, we use a simplified Transformer structure \cite{He24:ICLR}, with MSA and multi-layer perception (MLP) arranged in parallel, as shown in Fig. \ref{fig:RSSTE}(b). 
%MLP consists of two fully connected linear layers with a GELU layer in between.
%The outputs of MSA and MLP are combined through a final weighted sum to produce the final output of SSTE:
%\begin{equation}
%	\vspace{-0.1cm}
%	\label{eq:SSTE_OUTPUT}
%	F^{n}_{out}={\alpha}MSA(Norm(F^{n}_{in})) + {\beta}MLP(Norm(F^{n}_{in}))
%	\vspace{-0.1cm}
%\end{equation}
%According to \cite{Liang21:ICCVW}, we alternately use regular and shifted window partitioning across multiple SSTE layers to increase the correlation between partitioned windows.

We adopt a simplified Transformer structure \cite{He24:ICLR} for SSTE, where MSA and the multi-layer perceptron (MLP) are arranged in parallel, as shown in Fig. \ref{fig:RSSTE}(b). 
The MLP consists of two fully connected layers with a GELU activation in between. 
The outputs of MSA and MLP are combined via a weighted sum to produce the final SSTE output:
\begin{equation}
\vspace{-0.1cm}
\label{eq:SSTE_OUTPUT}
F^{n}_{out}={\alpha}MSA(Norm(F^{n}_{in})) + {\beta}MLP(Norm(F^{n}_{in}))
\vspace{-0.1cm}
\end{equation}
Based on the approach in \cite{Liang21:ICCVW}, we alternate between regular and shifted window partitioning across multiple SSTE layers to enhance the correlation between windows.

%Following \cite{Liang21:ICCVW}, we alternate between regular and shifted window partitioning across multiple SSTE layers to enhance the correlation between windows.

%\begin{figure*}[htbp]
%	\centering
%	\subfigure{
%		\includegraphics[width=6.9in]{Figure_comp_set8_tractor.png}
%		\label{fig:comp_set8_tractor}
%	}
%	%\vspace{0.1cm} % 控制上下图片间距
%	\subfigure{
%		\includegraphics[width=6.9in]{Figure_comp_DAVIS_giant_slalom.png}
%		\label{fig:comp_davis_slalom}
%	}
%	\caption{Video denoising comparison ($\sigma=50$) on Set8 and DAVIS}
%	\label{fig:comp_set8_davis}
%\end{figure*}

%\begin{figure*}[htbp]
%	\centering
%	\subfigure{
%		\includegraphics[width=6.6in]{Figure_comp_set8_tractor.png}
%		\label{fig:comp_set8_tractor}
%	}
%	%\vspace{0.1cm} % 控制上下图片间距
%	\subfigure{
%		\includegraphics[width=6.6in]{Figure_comp_DAVIS_giant_slalom.png}
%		\label{fig:comp_davis_slalom}
%	}
%	\caption{Video denoising comparison ($\sigma=50$) on Set8\cite{Tassano19:ICIP} and DAVIS\cite{Khoreva18:ACCV}.}
%	\label{fig:comp_set8_davis}
%\end{figure*}

%VLNB\cite{Arias18:JMIV}
%DVDNet\cite{Tassano19:ICIP}
%fastDVDNet\cite{Tassano20:CVPR}
%PaCNet\cite{Vaksman21:CVPR}
%RVRT\cite{Liang22:NIPS}

	\begin{table}[htbp]
	\vspace{-0.3cm}
	\caption{Average PSNR comparison on the Set8\cite{Tassano19:ICIP} and Davis\cite{Khoreva18:ACCV} dataset. Best in \textcolor{red}{\textbf{RED}} and second in \textcolor{blue}{\textbf{BLUE}}.}
	\vspace{-0.2cm}
	\centering
	\scalebox{1.0}{
		\hspace*{-0.4cm}
		\begin{tabular}{|c|c|c|c|c|c|c|}
			\hline
			DB & Method & $\sigma$=10 & $\sigma$=20 & $\sigma$=30 & $\sigma$=40 & $\sigma$=50 \\ \hline
			\multirow{6}{*}{Set8} & VLNB\cite{Arias18:JMIV} &37.26&33.72&31.74&30.39&29.24 \\ \cline{2-7}
			& DVDNet\cite{Tassano19:ICIP} &36.08&33.49&31.79&30.55&29.56 \\ \cline{2-7}
			& FastDVDNet\cite{Tassano20:CVPR} &36.44&33.43&31.68&30.46&29.53 \\ \cline{2-7}
			& PaCNet\cite{Vaksman21:CVPR} &37.06&33.94&32.05&30.70&29.66 \\ \cline{2-7}
			& RVRT\cite{Liang22:NIPS} &\textcolor{blue}{\textbf{37.53}}&\textcolor{blue}{\textbf{34.83}}&\textcolor{blue}{\textbf{33.30}}&\textcolor{red}{\textbf{32.21}}&\textcolor{red}{\textbf{31.33}} \\ \cline{2-7}
			& \textbf{GRTN}(ours) & \textcolor{red}{\textbf{37.62}} & \textcolor{red}{\textbf{34.96}} & \textcolor{red}{\textbf{33.37}} & \textcolor{blue}{\textbf{32.10}} & \textcolor{blue}{\textbf{31.22}} \\ \hline
			\multirow{6}{*}{Davis} & VLNB\cite{Arias18:JMIV} &38.85&35.68&33.73&32.32&31.13 \\ \cline{2-7}
			& DVDNet\cite{Tassano19:ICIP} &38.13&35.70&34.08&32.86&31.85 \\ \cline{2-7}
			& FastDVDNet\cite{Tassano20:CVPR} &38.71& 35.77 & 34.04 & 32.82 & 31.86 \\ \cline{2-7}
			& PaCNet\cite{Vaksman21:CVPR} &39.97& 36.82 & 34.79 & 33.34 & 32.20 \\ \cline{2-7}
			& RVRT\cite{Liang22:NIPS} &\textcolor{blue}{\textbf{40.57}}& \textcolor{blue}{\textbf{38.05}} & \textcolor{red}{\textbf{36.57}} & \textcolor{red}{\textbf{35.47}} & \textcolor{red}{\textbf{34.57}} \\ \cline{2-7}
			& \textbf{GRTN}(ours) & \textcolor{red}{\textbf{40.79}} & \textcolor{red}{\textbf{38.25}} & \textcolor{blue}{\textbf{36.56}} & \textcolor{blue}{\textbf{35.46}} & \textcolor{blue}{\textbf{34.47}} \\ \hline
		\end{tabular}
	}
	\label{tab:PSNR_SET8_DAVIS}
	%\vspace{-0.6cm}
\end{table}

\begin{table}[htbp]
	\vspace{-0.3cm}
	\caption{Average PSNR comparison on the Set8\cite{Tassano19:ICIP} for ablation study}
	\vspace{-0.2cm}
	\centering
	\scalebox{1.0}{
		\hspace*{-0.4cm}
		\begin{tabular}{|c|c|c|c|c|c|c|}
			\hline
			DB & Method & $\sigma$=10 & $\sigma$=20 & $\sigma$=30 & $\sigma$=40 & $\sigma$=50 \\ \hline
			\multirow{4}{*}{Set8} & Disable Gated scheme &36.52&34.36&32.95&31.79&30.96 \\ \cline{2-7}
			& Dot product attention&36.85&34.55&33.08&31.89&31.04 \\ \cline{2-7}
			& Disable Ortho loss &37.22&34.75&33.22&31.99&31.13 \\ \cline{2-7}
			& \textbf{GRTN} & 37.62 & 34.96 & 33.37 & 32.10 & 31.22 \\ \hline
		\end{tabular}
	}
	\label{tab:PSNR_Ablation}
	\vspace{-0.5cm}
\end{table}

\vspace{-0.05cm}
\section{Experiments}
\vspace{-0.15cm}

%	Based on the comparison of subjective and objective experimental results, our proposed network has achieved the same denoising performance as the SOTA multi-frame delay networks while maintaining a delay of just one frame.

% feature 96, total para: 2764779, spynet: 1440300, then it is 1324479
% feature 144, total para: 4221195, spynet: 1440300, then it is 2780895
% feature 192, total para: 6249003, spynet: 1440300, then it is 4808703‬

\subsection{Training}
\vspace{-0.05cm}

We configure each RSSTE with three SSTE layers, using a partition window size of $16{\times}16$, with 6 heads and 192 feature dimensions. 
The model contains a total of $4.81M$ parameters. 
The orthogonality loss weight $\lambda$ is set to 0.001.
Training is conducted on the DAVIS dataset\cite{Khoreva18:ACCV} with a patch size of $256{\times}256$ and a batch size of 8. 
Optimization follows the Cosine Annealing schedule \cite{Loshchilov17:ICLR}, starting with a learning rate of $4{\times}10^{-4}$ over 480,000 iterations. 
SpyNet \cite{Ranjan17:CVPR} is used to estimate video optical flow. 
Additive white Gaussian noise with noise level $\sigma \in [0, 50]$ is applied during training.
The network is implemented in PyTorch and trained on 8 NVIDIA A100 GPUs.

\vspace{-0.2cm}
\subsection{Comparison with SOTA Methods}
\vspace{-0.1cm}

We conduct video denoising experiments on the DAVIS\cite{Khoreva18:ACCV} and SET8\cite{Tassano19:ICIP} test datasets, using Gaussian noise at levels of $[10, 20, 30, 40, 50]$. 
To ensure a fair comparison with SOTA multi-frame-delay methods, 
we duplicate the first 16 frames of each test scene and prepend them to the first frame. 
These replicated frames are excluded from PSNR calculations.
The PSNR comparisons between our GRTN and other SOTA methods are shown in Table \ref{tab:PSNR_SET8_DAVIS}. 
GRTN achieves comparable denoising performance to RVRT, with only a single-frame delay, and outperforms RVRT when noise level $\sigma < 30$.
Visual comparisons in Fig. \ref{fig:comp_set8_davis} further confirm that GRTN performs on par with SOTA methods.

\vspace{-0.2cm}
\subsection{Ablation Study}
\vspace{-0.1cm}

%To validate the effectiveness of each module, we perform an ablation study on the Set8 dataset, with PSNR comparisons presented in Table 2.
%
%To evaluate the impact of the gated scheme, we observe a PSNR decrease of 0.3 compared to GRTN.
%
%To assess the effect of using Euclidean distance in the Transformer, we find a PSNR drop of approximately 0.2 compared to GRTN with Euclidean distance-based attention.
%
%Lastly, to examine the influence of orthogonality loss, we observe a PSNR reduction of about 0.1 compared to GRTN.

%To validate the effectiveness of each module, we conduct an ablation study on the Set8 dataset\cite{Tassano19:ICIP}, with PSNR comparisons shown in Table \ref{tab:PSNR_Ablation}.
%To evaluate the impact of the gated scheme, we disable the reset gate, update gate, and blending mechanisms, 
%while maintaining the same training method and parameters as GRTN. 
%This results in a PSNR decrease of approximately 0.26dB for noise $\sigma=50$ compared to GRTN.
%To assess the effect of using Euclidean distance in the Transformer, we replace the Euclidean distance-based self-attention in RSSTE with standard dot product-based self-attention and retrain the model.
%We find a PSNR drop of approximately 0.18dB for noise $\sigma=50$ compared to GRTN with Euclidean distance-based attention.
%Lastly, to examine the influence of orthogonality loss, we set the weight $\lambda$ to 0 and retrain the model. 
%We observe a PSNR reduction of about 0.09dB for noise $\sigma=50$ compared to GRTN.

To validate the effectiveness of each module, we conduct an ablation study on the Set8 dataset\cite{Tassano19:ICIP}, with PSNR comparisons shown in Table \ref{tab:PSNR_Ablation}.
To evaluate the impact of the gated scheme, we disable the reset gate, update gate, and blending mechanisms, 
while maintaining the same training method and parameters as GRTN. 
This results in a PSNR decrease of approximately 0.26dB for noise $\sigma=50$.
To assess the effect of using Euclidean distance in the Transformer, we replace the Euclidean distance-based self-attention in RSSTE with standard dot product-based self-attention and retrain the model.
We find a PSNR drop of approximately 0.18dB for noise $\sigma=50$.
Lastly, to examine the influence of orthogonality loss, we set the weight $\lambda$ to 0 and retrain the model. 
We observe a PSNR reduction of about 0.09dB for noise $\sigma=50$.

\vspace{-0.1cm}
\section{Conclusion}
\vspace{-0.1cm}

%In this paper, to address the issue that current state-of-the-art video denoising methods cannot meet the stringent requirement of 1-frame delay for practical imaging devices, we propose the Gated Recurrent Transformer Network (GRTN). Our GRTN achieves video denoising quality comparable to that of state-of-the-art methods with multi-frame delay (e.g., 16 frames) while significantly reducing complexity.
%Experimental results validate the effectiveness and superiority of our proposed GRTN method. In future work, we plan to extend the GRTN network to other computer vision problems with stringent delay requirements.

%In this paper, we propose GRTN to overcome the limitations of current SOTA video denoising methods, which cannot meet the strict 1-frame delay requirement for practical imaging devices. GRTN achieves comparable denoising quality to multi-frame delay SOTA methods (e.g., 16 frames) with only a 1-frame delay. Experimental results demonstrate the effectiveness and superiority of GRTN. In future work, we plan to extend the GRTN framework to other computer vision tasks with stringent delay constraints.

In this paper, we propose GRTN, which achieves denoising quality comparable to multi-frame delay SOTA methods (e.g., 16 frames) with only a single-frame delay. 
Experimental results demonstrate the effectiveness and superiority of GRTN. 
In future work, we plan to extend GRTN to other computer vision tasks with strict delay constraints.

\end{document}